\documentclass[journal]{IEEEtran}

\usepackage{amsmath,amssymb}
\usepackage{booktabs}
\usepackage{graphicx}
\usepackage{url}
\usepackage{cite}

\newcommand{\Rel}{\mathbf{o}}
\newcommand{\dO}{\Delta\Rel}
\newcommand{\gradV}{\nabla V}
\newcommand{\DGR}{\mathrm{DGR}}

\begin{document}

\title{Beyond Perturbation Magnitude: Direction-Dependent Responses in Multimodal Geometric Representations}

\author{Yongsheng~Luo, Wengan~He, Yu~Li, Rouying~Wu, and Wei~Lv%
\thanks{This work was supported by the Guangdong Key Disciplines Project under Grant 2024ZDJS137.}%
\thanks{Y. Luo is with the Zhuhai College of Science and Technology, Zhuhai, China (e-mail: vincent@zcst.edu.cn).}%
\thanks{W. He is with the Zhuhai College of Science and Technology, Zhuhai, China, and also with Zhuhai UNO Technology Co., Ltd., Zhuhai, China (e-mail: hewengan@essex.com.cn).}%
\thanks{Y. Li is with the Zhuhai College of Science and Technology, Zhuhai, China (e-mail: jluzhliyu@zcst.edu.cn).}%
\thanks{R. Wu is with the Zhuhai College of Science and Technology, Zhuhai, China (e-mail: wurouying@outlook.com).}%
\thanks{W. Lv is with the Zhuhai College of Science and Technology, Zhuhai, China (e-mail: luwei@zcst.edu.cn). (Corresponding author: Wei Lv.)}}
\maketitle

\begin{abstract}
Geometric alignment scores built from Gram determinants model higher-order consistency among modalities, yet how they respond to modality degradation is poorly understood. We ask whether this response is organized by the magnitude of the displacement that degradation induces. Using frozen cohorts from MSR-VTT (N=878), DiDeMo (N=980), and Charades (N=1861), we degrade one modality at a time with controlled video blur and audio noise, and measure the displacement in the relational geometry on which the score is defined. Displacement magnitude explains a substantial share of the pooled variance of the absolute volume response (out-of-sample coefficient of determination 0.47--0.70 across 21 conditions), but within narrow magnitude bins it ranks the larger response at chance (0.50--0.53). The closed-form first-order expansion of the Gramian volume yields a directional term: the projection of the displacement onto the local volume gradient. Across all 21 conditions this term explains 0.84--0.998 of the pooled variance, ranks within-magnitude responses with accuracy 0.84--0.99, and recovers the response sign with accuracy 0.91--0.997. Direction-free combinations of local gain and magnitude, and encoder-side sensitivity signals, fall clearly short. A flexible learned regressor is comparable overall and better on audio; its advantage is a smooth function of the observed displacement beyond first order. A pre-specified gain-normalization candidate fails its gates. Because the term uses the observed degraded displacement, it is explanatory rather than a deployment-time predictor, and a clean-side approximation predicts audio responses partly but not video responses.
\end{abstract}


\section{Introduction}
\label{sec:intro}

\IEEEPARstart{M}{ultimodal} representation learning maps heterogeneous signals into shared embedding spaces, where agreement across modalities is measured by pairwise similarity. This design underlies contrastive vision--language models \cite{radford2021clip,jia2021align}, joint video--audio--text embeddings \cite{akbari2021vatt,girdhar2023imagebind,zhu2024languagebind,chen2023vast}, and video--text retrieval built on such representations \cite{luo2022clip4clip}. Beyond pairwise similarity, recent work defines geometric scores over several modality embeddings jointly: GRAM measures the volume of a Gramian parallelotope as a higher-order alignment quantity \cite{cicchetti2025gram}, and HyperGRAM extends the construction with Euclidean--hyperbolic geometry \cite{na2026hypergram}. Because the geometry is part of the learning objective, the behavior of the geometric score under degraded inputs affects how a model treats difficult or corrupted samples.

This raises a basic question for interpreting such scores: \emph{how does the geometric score respond when one modality is degraded?} A natural first hypothesis is magnitude-based: a stronger perturbation should produce a stronger change, and the displacement magnitude should organize the response. Related reliability-aware and incomplete-multimodal work \cite{wang2020gradientblending,ma2021smil,han2021trusted,wang2023shaspec,zhang2023provable} suggests that a scalar quality signal may summarize how strongly a degraded sample is affected, and corruption studies of unimodal models analyze behavior as a function of corruption strength \cite{hendrycks2019corruptions}.

We test this hypothesis directly.\footnote{A preliminary version of this work was posted as arXiv:2610.08533~\cite{luo2026arxiv} (versions v1 and v2). The present manuscript is not identical to that preprint: it corrects the displacement used for the magnitude result, extends the analysis to all severities and to a third dataset, and changes several conclusions. The differences are described in the statement submitted with the manuscript.} We freeze the representation and construct controlled degradations on three video--text--audio benchmarks, MSR-VTT \cite{xu2016msrvtt}, DiDeMo \cite{hendricks2017didemo}, and Charades \cite{sigurdsson2016charades}. For each sample we measure the displacement that degradation induces in the relational geometry, namely the vector of pairwise cosine similarities on which the score is defined, and compare it with the change in the volume. The answer has two parts. Magnitude explains a large share of the pooled variance of the response, because the response scales with the displacement. Yet magnitude does not organize the response \emph{within} a magnitude: samples displaced by the same amount respond differently, and magnitude ranks their responses at chance.

The missing structure follows from the geometry of the score. Let $\Rel=(C_{TV},C_{TA},C_{VA})$ collect the three pairwise cosine similarities among text, video, and audio embeddings. The Gramian volume $V(\Rel)$ has a closed-form gradient, so the first-order response to a displacement $\dO=\Rel_d-\Rel_c$ from a clean operating point $\Rel_c$ is
\begin{equation}
\begin{aligned}
\Delta V &\approx \gradV(\Rel_c)^{\top}\dO\\
&= \underbrace{\|\gradV(\Rel_c)\|}_{\text{local gain}}\,
\underbrace{\|\dO\|}_{\text{magnitude}}\,
\underbrace{\cos\theta}_{\text{direction}},
\end{aligned}
\label{eq:intro-decomp}
\end{equation}
where $\theta$ is the angle between the displacement and the local gradient. We call $\DGR(\Rel_c,\dO)=\gradV(\Rel_c)^{\top}\dO$ the \emph{Directional Geometric Response}. Since it uses the observed degraded displacement, it is an explanatory quantity, not a deployment-time predictor.

\begin{figure*}[t]
\centering
\includegraphics[width=\textwidth]{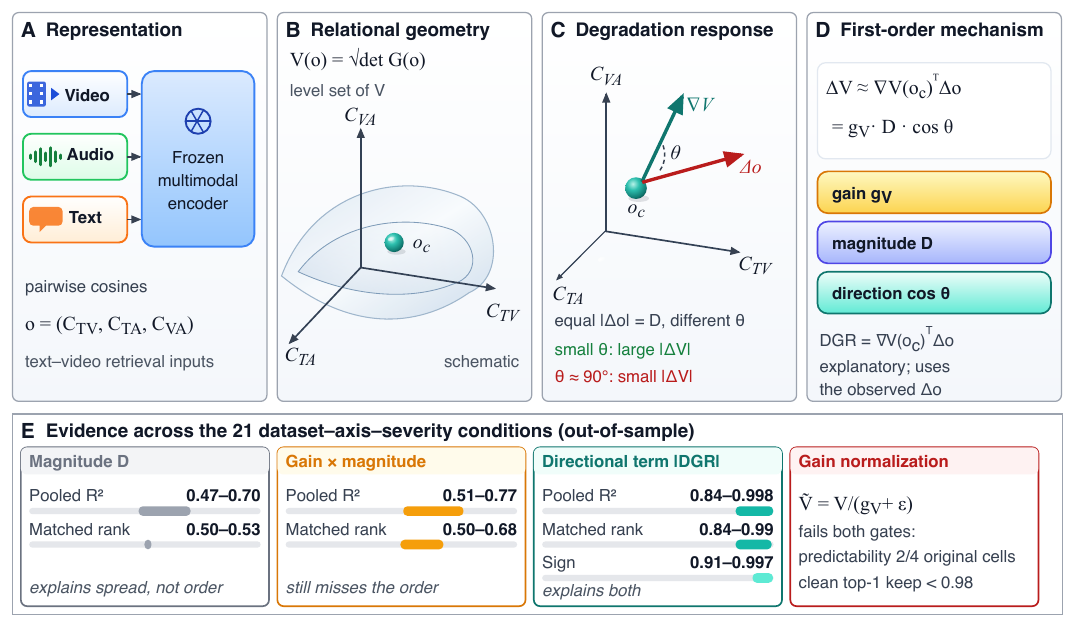}
\caption{Mechanism overview. (A)~A frozen encoder maps text, video, and audio to embeddings; their pairwise cosines define the relational coordinate $\Rel$. (B)~The Gramian volume $V(\Rel)$ over the relational coordinate (schematic). (C)~A degradation moves the operating point by $\dO$; equal-magnitude displacements in different directions produce different responses (schematic). (D)~The first-order response factors into local gain, magnitude, and direction; DGR uses the observed $\dO$ and is explanatory, not a predictor. (E)~Evidence from the 21 dataset--axis--severity conditions: magnitude explains the pooled spread but not the order within a magnitude bin, direction-free products remain incomplete, the directional term explains both, and gain normalization fails its pre-specified gates. Ranges are out-of-sample values from Tables I--II and Section~V.}
\label{fig:overview}
\end{figure*}

Fig.~\ref{fig:overview} summarizes the mechanism and the evidence behind it.

The paper makes five contributions, stated as empirical findings rather than new modules.
\begin{itemize}
\item We separate the roles of magnitude. Across three datasets, two perturbation families, and all tested severities, displacement magnitude explains a substantial share of the pooled response variance, but magnitude-matched samples are ordered at chance.
\item We show that the directional first-order term explains the within-magnitude ordering and the sign of the response in all 21 conditions, and that it outperforms direction-free combinations of local gain and magnitude, and encoder-side sensitivity signals, under the same evaluation instruments. A flexible learned regressor is higher on audio. An ablation and a residual analysis show that its advantage is a smooth function of the observed displacement beyond first order, not a clean-state effect.
\item We report a pre-specified gain-normalization audit. The tested family $V/(g_V+\varepsilon)$ does not consistently improve predictability and does not preserve clean ordering, so we do not turn the mechanism into a score-side calibration method.
\item We test the consequence for text-to-video retrieval. Under gallery degradation, the directional term accounts for which top-ranked order relations reverse, beyond what the clean margin and direction-free alternatives explain.
\item We characterize the limits of the mechanism: the first-order approximation is accurate for video but has an 18--41\% median relative error for audio at the tested strengths, and a clean-side approximation of the response is informative for audio but not for video.
\end{itemize}

The scope is deliberately bounded. The evidence covers three datasets and two perturbation families, one Gramian-volume geometry, and no causal or universal claim. The first-order expansion is elementary, and we do not present it as a mathematical contribution; the contribution is the empirical separation of magnitude and direction, and the falsification-oriented evaluation that supports it.

The remainder of the paper is organized as follows. Section~\ref{sec:related} positions the study. Section~\ref{sec:setup} gives the geometry, the perturbation protocol, and the evaluation instruments. Sections~\ref{sec:anomaly}--\ref{sec:mechanism} report the magnitude analysis, the falsification of simpler explanations, and the directional mechanism. Section~\ref{sec:clean} examines clean-side approximations, Section~\ref{sec:boundary} reports the gain-normalization audit, and Section~\ref{sec:discussion} discusses implications and limitations.

\section{Related Work}
\label{sec:related}

\subsection{Multimodal Representation and Geometric Alignment}
Shared embedding spaces are a standard mechanism for multimodal learning \cite{baltrusaitis2019multimodal}. CLIP establishes image--text alignment through contrastive learning \cite{radford2021clip}, ALIGN scales the recipe with noisy web-scale text \cite{jia2021align}, and VATT extends it to raw video, audio, and text \cite{akbari2021vatt}. ImageBind and LanguageBind connect a larger number of modalities in one space \cite{girdhar2023imagebind,zhu2024languagebind}, and VAST models vision, audio, subtitle, and text jointly \cite{chen2023vast}. CLIP4Clip studies the transfer of contrastive image--text representations to video--text retrieval \cite{luo2022clip4clip}.

Joint geometric constructions replace purely pairwise alignment. GRAM measures a Gramian volume over multiple modality embeddings \cite{cicchetti2025gram}, and HyperGRAM extends it to hyperbolic space \cite{na2026hypergram}. Hyperbolic representation learning has been motivated by its capacity to represent hierarchies \cite{nickel2017poincare,ganea2018hyperbolic,peng2022hyperbolic}, and has been applied to image--text learning \cite{desai2023meru,poppi2025hyperbolic}. Metric-geometry considerations are studied in \cite{musgrave2020metric}. These works mainly evaluate representation quality and retrieval. We hold the representation fixed and ask how its geometric score itself changes under controlled modality perturbation.

\subsection{Missing Modalities, Perturbation, and Reliability}
Modalities differ in quality, availability, and optimization behavior \cite{baltrusaitis2019multimodal,wang2020gradientblending}. Methods address severely missing modalities \cite{ma2021smil,wang2023shaspec,lee2023prompting}, quality-aware fusion \cite{zhang2023provable}, and sample-dependent evidence for multiple views \cite{han2021trusted}. These works motivate the hypothesis that a scalar quality or reliability variable could describe the response of a geometric score under degradation. Our setting differs: we examine the local response of an already-trained score rather than optimize a model to compensate for unreliable modalities. A reliability variable can describe \emph{how much} a modality is compromised, but not necessarily \emph{which relational direction} the resulting displacement takes. Our results show that the latter matters for the score studied here.

Gradient-based sensitivity is a standard tool: saliency methods \cite{simonyan2014saliency}, influence functions \cite{koh2017influence}, and adversarial-perturbation studies \cite{szegedy2014intriguing,goodfellow2015explaining,moosavi2016deepfool,madry2018towards} show that small input changes can produce nonuniform responses. This paper is not an adversarial-robustness study; its perturbations are controlled signal degradations of a frozen encoder. Because the volume has a closed-form derivative, the first-order response can be tested directly against the measured change, and the direction of the induced displacement in relational coordinates becomes observable. What remains underexplored is how a geometric multimodal score responds to the \emph{direction} of the displacement induced in its own relational geometry, which this paper addresses.

\section{Geometric Formulation and Evaluation Protocol}
\label{sec:setup}

\subsection{Multimodal Relational Geometry}
Let $t,v,a\in\mathbb{R}^m$ be the $\ell_2$-normalized text, video, and audio embeddings, with pairwise cosine similarities $x=t^\top v$, $y=t^\top a$, $z=v^\top a$, and relational coordinate $\Rel=(x,y,z)$. The Gram matrix is
\begin{equation}
\begin{aligned}
G(\Rel)&=
\begin{bmatrix}
1&x&y\\
x&1&z\\
y&z&1
\end{bmatrix},\\
d(\Rel)&=\det G(\Rel)=1+2xyz-x^2-y^2-z^2.
\end{aligned}
\label{eq:det}
\end{equation}
The analyzed score is the Gramian volume component of the deployment scoring path, $V(\Rel)=\sqrt{|\det G|}$. In the cohorts used here all determinants are nonnegative, so $V=\sqrt{d}$ is differentiable at every observed clean point. The deployed HyperGRAM score also contains a hyperbolic component and an interaction re-ranking stage; we analyze the Euclidean component as implemented. Used alone for text-to-video ranking on the analyzed cohort, this component places the correct video first for $42.9$\% of queries (Section~\ref{sec:retrieval}).

\subsection{Perturbation Response and Displacement Magnitude}
Let $\Rel_c$ and $\Rel_d$ be the relational coordinates of a sample before and after one modality is degraded. The response and the displacement magnitude are
\begin{equation}
\Delta V=V(\Rel_d)-V(\Rel_c),
\qquad
D=\|\dO\|=\|\Rel_d-\Rel_c\|.
\label{eq:response}
\end{equation}
$D$ is the Euclidean length of the displacement in the relational coordinates after the frozen encoders and the pairwise cosine construction. It is not a raw pixel, waveform, or embedding perturbation strength. Two degradations of very different raw severity can produce similar $D$, and the relational coordinate absorbs both.

\subsection{Directional Geometric Response}
\begin{figure}[t]
\centering
\includegraphics[width=0.78\columnwidth]{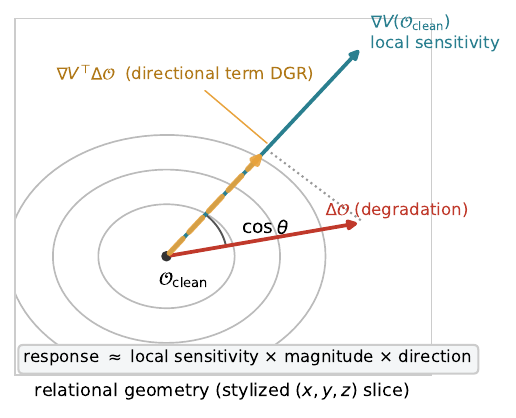}
\caption{Local decomposition of the first-order response. A degradation moves the relational coordinate by $\dO$ from $\Rel_c$; the induced volume change is the projection of $\dO$ onto $\gradV(\Rel_c)$, i.e., gain $\times$ magnitude $\times\cos\theta$.}
\label{fig:schematic}
\end{figure}

The determinant gradient is $J_d(\Rel)=(2yz-2x,\;2xz-2y,\;2xy-2z)$, and since $V=\sqrt{d}$,
\begin{equation}
\gradV(\Rel)=\frac{J_d(\Rel)}{2V(\Rel)}.
\label{eq:gradient}
\end{equation}
A first-order expansion gives $\Delta V\approx \gradV(\Rel_c)^{\top}\dO$, and we define
\begin{equation}
\begin{aligned}
\DGR(\Rel_c,\dO)&=\gradV(\Rel_c)^{\top}\dO,\\
g_V(\Rel_c)&=\|\gradV(\Rel_c)\|.
\end{aligned}
\label{eq:dgr}
\end{equation}
so that $\DGR=g_V\,D\cos\theta$. Three factors enter the first-order response: the local gain $g_V$ at the operating point, the magnitude $D$, and the alignment $\cos\theta$ between the displacement and the local gradient. Fig.~\ref{fig:schematic} illustrates the decomposition. The expansion is an approximation, not an identity; whether it explains observed responses, and whether the direction factor is necessary, are empirical questions addressed below. $\DGR$ requires the degraded displacement $\dO$, so it is an explanatory quantity and not a deployment-time predictor, reliability estimator, or calibration variable.

\subsection{Datasets and Controlled Degradations}
\label{sec:datasets}
We use MSR-VTT \cite{xu2016msrvtt} and DiDeMo \cite{hendricks2017didemo}, video--text benchmarks with text, video, and audio. The MSR-VTT cohort contains $N=878$ videos after excluding 116 videos without an audio track and 6 silent tracks; membership is fixed by an identifier list. The DiDeMo cohort contains $N=980$ videos after excluding 5 videos without audio and 19 silent tracks. The Charades cohort uses the 1863 test-split videos of Charades \cite{sigurdsson2016charades}, each described by its free-text script; all have an audio track, and two with near-zero audio power are excluded, giving $N=1861$. Charades audio is much quieter than the other two datasets and often sparse, and its clips are longer (median 31~s); Supplementary Section~S13 documents these properties, including the difference between the clip-level SNR used here and the level seen by the encoder. ActivityNet \cite{cabahelbron2015activitynet} would provide an external test and is left for future work.

One modality is degraded at a time and the others are kept clean. On the video axis we apply Gaussian blur with $\sigma\in\{1,2,4\}$; on the audio axis we add white Gaussian noise at $\mathrm{SNR}\in\{20,10,5,0\}$~dB. Text is always clean. A \emph{cell} is one dataset paired with one degradation axis. Each cell is analyzed at every severity level, and the four strongest levels ($\sigma=4$ or $\mathrm{SNR}=0$~dB) serve as the headline setting. This yields twenty-one dataset--axis--severity conditions, seven of them on Charades.

\subsection{Evaluation Instruments}
\label{sec:protocol}
All analyses use the same three instruments.

\emph{Pooled out-of-sample coefficient of determination.} For a candidate $s$ we fit an ordinary least-squares model of $|\Delta V|$ on $s$ with an intercept and report the pooled five-fold out-of-sample $R^2$ over all held-out predictions. Within each cell and severity, one fold partition (seed 3105) is shared by all candidates, so comparisons reflect the candidates rather than the partition. The closed-form fit has no hyperparameters and nothing is tuned.

\emph{Matched-magnitude ranking.} For each pair of samples whose displacements are close, a candidate receives credit when it ranks the sample with the larger $|\Delta V|$ correctly; chance is $0.5$. A candidate that is determined by magnitude alone scores at chance, and a candidate must recover the ordering within a magnitude bin. The primary window is relative: pairs satisfy $|\log D_i-\log D_j|<0.05$. Because the relational displacement varies by roughly an order of magnitude across severities, we found an absolute window ($|D_i-D_j|<0.01$, used in an earlier internal draft) not scale-invariant; it is reported as a sensitivity check in the supplementary material, and the change is disclosed here.

\emph{Bootstrap intervals.} Confidence intervals are sample-level (cluster) bootstrap intervals with 300 resamples: samples are resampled with replacement, matched pairs are rebuilt within each resample, and paired comparisons between two candidates use the same resample \cite{efron1993bootstrap,cameron2015cluster}. Reproducibility conventions follow \cite{pineau2021reproducibility}.

\emph{Order of analysis.} The directional term was identified in exploratory analysis of the same cohorts. The proxy candidates of Section~\ref{sec:proxies} were specified before they were evaluated, and the decision rules of the gain-normalization audit (Section~\ref{sec:boundary}) were fixed before that audit was run. The four-cell, all-severity analysis reported here is a confirmatory replication of the exploratory result, and we have not altered any decision rule after observing outcomes.

\section{Magnitude Explains Variance, Not Ordering}
\label{sec:anomaly}

\begin{figure*}[t]
\centering
\includegraphics[width=\textwidth]{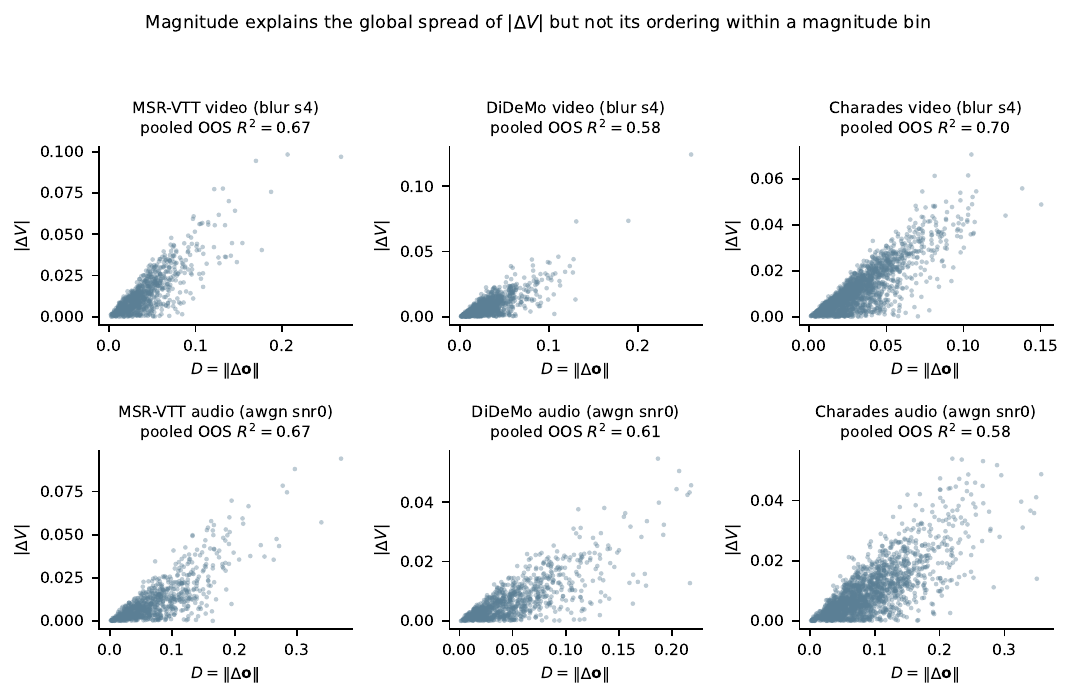}
\caption{Absolute volume response $|\Delta V|$ against relational displacement $D=\|\dO\|$ at the strongest level of each cell (top: video blur $\sigma=4$; bottom: audio noise SNR$=0$~dB; columns: MSR-VTT ($N=878$), DiDeMo ($N=980$), Charades ($N=1861$)). Magnitude accounts for the global spread but leaves the ordering of samples within a magnitude bin largely unexplained.}
\label{fig:magnitude}
\end{figure*}

If magnitude organized the response, $D$ would explain $|\Delta V|$ both globally and within magnitude bins. The first part holds: at every condition $D$ explains between $0.47$ and $0.70$ of the pooled out-of-sample variance of $|\Delta V|$ (Table~\ref{tab:ladder}). This reflects that the response grows with the displacement. The second part does not hold. Within magnitude bins, the ranking accuracy of $D$ is $0.50$--$0.51$ at the strongest level in all six cells, and between $0.50$ and $0.53$ across all severities (Fig.~\ref{fig:magnitude}). Samples displaced by the same amount can respond differently in size and in sign. At the strongest video blur the mean displacement is $\bar D\approx0.042$ and 403 of 878 samples respond negatively; at the strongest audio noise the mean is $\bar D\approx0.121$ and 211 of 878 respond negatively. A magnitude-only account cannot express this heterogeneity.

\section{Simpler Explanations Fail}
\label{sec:proxies}

We evaluate the natural scalar alternatives with the same instruments. Encoder-side and operating-point signals were computed on MSR-VTT only, where the proxy protocol was run.

\emph{Encoder sensitivity} (S2) is the change of the unimodal encoder under a fixed mild probe, and \emph{self-consistency} (S3) is the disagreement between two benign views of the same clean input. Both encode notions of sample vulnerability. Neither orders matched-magnitude responses above chance on either MSR-VTT axis (ranking accuracy $0.47$--$0.49$), and neither adds more than $0.007$ to the pooled $R^2$.

\emph{Clean operating point} (S1), that is the clean Gramian volume, ranks matched responses below chance on video ($0.38$) and near chance on audio ($0.45$).

\emph{Direction-free clean gain.} The clean-side gradient norm $\|J_d(\Rel_c)\|$ is the strongest direction-free signal we found. On the MSR-VTT video axis at $\sigma=4$ it orders matched responses with accuracy $0.65$ (cluster interval $[0.63,0.68]$) and adds $0.054$ to the pooled $R^2$ of $D$ (interval $[0.040,0.070]$). On audio it is at chance ($0.50$). This is a genuine but partial signal: it passes the matched-ranking gate on video only, and it does not explain the response on audio.

\emph{Direction-free product.} The product $\|\gradV\|\cdot D$ combines the clean gain with magnitude but discards direction. It improves over $D$ in pooled explanation ($R^2$ $0.56$--$0.77$ versus $0.47$--$0.67$), yet remains well below the directional term (Section~\ref{sec:mechanism}). The first-order decomposition explains why: the product keeps gain and magnitude but drops $\cos\theta$, and the direction of the displacement varies across samples.

\begin{table*}[t]
\centering
\caption{Falsification ladder at the strongest level of each cell. Pooled out-of-sample $R^2$ and matched-magnitude ranking accuracy (relative window; chance $0.5$), listed in the order MSR-VTT video, MSR-VTT audio, DiDeMo video, DiDeMo audio; the Charades cells are reported in Table~\ref{tab:main}. Encoder-side and clean-operating-point rows are computed on MSR-VTT only ($\dagger$); for them the entry is the incremental $R^2$ over $D$ on the video and audio axes.}
\label{tab:ladder}
\small
\begin{tabular*}{\textwidth}{@{\extracolsep{\fill}}p{2.6cm}p{4.6cm}p{3.6cm}p{3.6cm}l@{}}
\toprule
Candidate & Information used & Pooled $R^2$ & Matched rank & Decision\\
\midrule
$D$ & displacement magnitude & 0.67, 0.67, 0.58, 0.61 & 0.51, 0.51, 0.50, 0.50 & Rejected\\
S2$^\dagger$ & encoder sensitivity & $+0.005$, $+0.001$ & 0.47, 0.48 & Rejected\\
S3$^\dagger$ & self-consistency & $+0.007$, $+0.005$ & 0.47, 0.49 & Rejected\\
S1$^\dagger$ & clean volume & -- & 0.38, 0.45 & Rejected\\
$\|J_d\|^\dagger$ & clean gain alone & $+0.054$, $0.000$ & 0.65, 0.50 & Video only\\
$\|\gradV\|D$ & gain $\times$ magnitude & 0.77, 0.63, 0.69, 0.59 & 0.65, 0.51, 0.66, 0.55 & Incomplete\\
$|\DGR|$ & gain $\times$ magnitude $\times$ direction & \textbf{0.97, 0.92, 0.94, 0.84} & \textbf{0.96, 0.88, 0.96, 0.84} & Supported\\
\bottomrule
\end{tabular*}
\end{table*}

\section{The Directional Geometric Response}
\label{sec:mechanism}

The first-order expansion gives $\Delta V=\DGR+O(\|\dO\|^2)$ with $\DGR$ as in \eqref{eq:dgr}. Table~\ref{tab:ladder} reports that $|\DGR|$ is the only candidate that explains both the pooled variance and the within-magnitude ordering at all six strongest cells, and Table~\ref{tab:main} gives the full comparison.

\begin{table*}[t]
\centering
\caption{Main results at the strongest level of each cell. Columns: out-of-sample $R^2$ for $|\Delta V|$ of $D$, $\|\gradV\|D$, and $|\DGR|$; the difference $|\DGR|$ minus $\|\gradV\|D$ with its sample-level bootstrap interval; matched-magnitude ranking accuracy of $|\DGR|$; sign accuracy of $\DGR$.}
\label{tab:main}
\small
\begin{tabular}{@{}lccccc@{}}
\toprule
Cell & $D$ & $\|\gradV\|D$ & $|\DGR|$ & $|\DGR|-\|\gradV\|D$ [95\% CI] & Sign\\
\midrule
MSR-VTT video ($N$=878) & 0.674 & 0.768 & \textbf{0.966} & 0.198 [0.154, 0.264] & 0.990\\
MSR-VTT audio ($N$=878) & 0.674 & 0.629 & \textbf{0.924} & 0.295 [0.248, 0.351] & 0.940\\
DiDeMo video ($N$=980) & 0.582 & 0.688 & \textbf{0.935} & 0.247 [0.156, 0.405] & 0.986\\
DiDeMo audio ($N$=980) & 0.606 & 0.590 & \textbf{0.836} & 0.246 [0.202, 0.293] & 0.909\\
Charades video ($N$=1861) & 0.697 & 0.762 & \textbf{0.986} & 0.224 [0.192, 0.261] & 0.996\\
Charades audio ($N$=1861) & 0.579 & 0.537 & \textbf{0.868} & 0.331 [0.297, 0.369] & 0.932\\
\bottomrule
\end{tabular}
\end{table*}

\begin{figure*}[t]
\centering
\includegraphics[width=\textwidth]{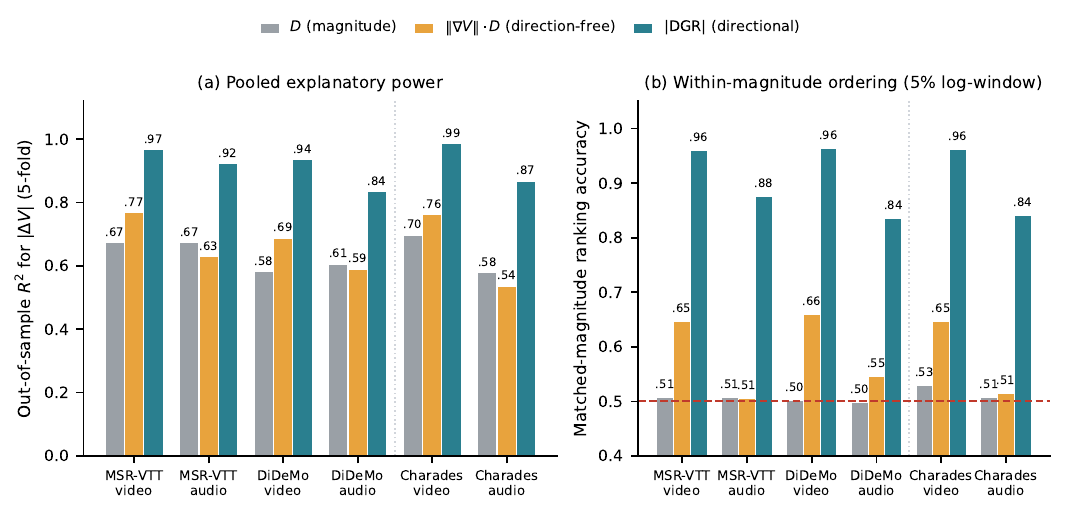}
\caption{Explanatory ladder at the strongest level of each of the six cells. (a) Pooled out-of-sample $R^2$ for $|\Delta V|$ under the magnitude $D$, the direction-free product $\|\gradV\|D$, and the directional term $|\DGR|$. (b) Matched-magnitude ranking accuracy (relative window; chance $0.5$). Magnitude and direction-free products explain the global spread partly, but only the directional term orders responses within a magnitude bin.}
\label{fig:ladder}
\end{figure*}

Across the 21 conditions, $|\DGR|$ explains $0.84$--$0.998$ of pooled out-of-sample variance, against $0.51$--$0.77$ for $\|\gradV\|D$, and each cell's difference from $\|\gradV\|D$ has an interval whose lower bound is at least $0.15$ (Supplementary Table~S1). Its matched-magnitude ranking accuracy is $0.84$--$0.99$, against $0.50$--$0.53$ for $D$ at the strongest levels and $0.50$--$0.68$ for $\|\gradV\|D$. The signed term recovers the sign of the response with accuracy $0.91$--$0.997$ across all conditions. Fig.~\ref{fig:ladder} shows all six cells at the strongest levels.

\subsection{First-Order Remainder and Its Bound}
The approximation is not uniformly exact, and we bound its error rather than rely on the observed deviations alone. Taylor's theorem with Lagrange remainder gives $\Delta V-\DGR=\tfrac12\,\dO^{\top}\nabla^2V(\xi)\,\dO$ for some $\xi$ on the segment from $\Rel_c$ to $\Rel_d$, where $\nabla^2V=H_d/(2V)-J_dJ_d^{\top}/(4V^{3})$ and $H_d$ is the Hessian of $d$. Hence
\begin{equation}
|\Delta V-\DGR|\le B:=\tfrac12\sup_{t\in[0,1]}\bigl|\dO^{\top}\nabla^2V(\Rel_c+t\,\dO)\,\dO\bigr|.
\label{eq:bound}
\end{equation}
We evaluate $B$ in \eqref{eq:bound} for every sample on a 401-point grid; the segment stays at $V\ge0.72$ in all cells. The bound holds for all $5{,}577$ strongest-level samples across the six cells (3{,}716 original and 1{,}861 Charades), and it is nearly attained: the median ratio $B/|\Delta V-\DGR|$ is $1.01$--$1.02$. Since $|\DGR|=g_V D|\cos\theta|$, the relative error is controlled by $D\sup\|\nabla^2V\|/(2g_V|\cos\theta|)$. The median Hessian norm is similar across cells ($1.6$--$1.9$), so the audio--video gap comes mainly from displacement size: the median $D$ is about $0.08$--$0.12$ on audio and $0.03$--$0.04$ on video. The bound ranks per-sample relative errors (Spearman $0.51$--$0.72$ on the original cells), but it does not reproduce the full gap between axes, which also depends on the direction factor. At the strongest levels the median relative error is $6.0$--$7.4$\% on video and $37.8$--$40.7$\% on audio across the six cells; across all severities it is $2.5$--$7.4$\% and $18$--$41$\%. The correct reading is a bounded first-order mechanism, not an exact law. Supplementary Table~S4 gives the per-cell bounds.

\subsection{Comparison with a Learned Regressor}
A flexible learner with the same observed information gives a reference point for the first-order term. We trained a gradient-boosted regressor (default hyper-parameters, no tuning) on the clean coordinates and Gram volume $(C_{TV},C_{TA},C_{VA},V_{\rm clean})$ and on the observed displacement $(\Delta C_{TV},\Delta C_{TA},\Delta C_{VA})$, under the same five folds. Its pooled out-of-sample $R^2$ for $|\Delta V|$ is $0.68$--$0.97$ across the 21 conditions, against $0.84$--$1.00$ for the one-parameter recalibration of $|\DGR|$. The directional term is higher in 13 conditions: all nine video conditions by $0.02$--$0.27$, and four audio conditions by up to $0.03$. The learner is higher in the other eight audio conditions, by $0.004$--$0.09$; the largest gaps occur on Charades and DiDeMo at low SNR. A ridge regression on the clean state and the per-coordinate products reaches $0.81$--$0.96$ on the original cells. The clean state alone predicts no video response, and it is informative for audio, most strongly at low SNR.

Two analyses explain the audio gap (Supplementary Tables~S6 and S7). First, an ablation: a learner given only the observed displacement reaches $0.60$--$0.74$ on audio, because it lacks the local gain. Adding $|\DGR|$ as a single feature to the displacement recovers the full learner performance (for example $0.96$ on Charades at SNR $0$~dB, against $0.96$ for the full seven-input learner and $0.86$ for $|\DGR|$ alone). Second, a residual test: after subtracting the first-order term, the remainder is predictable from the observed displacement alone, with out-of-sample $R^2$ of $0.81$--$0.96$ in all twelve audio conditions, and adding the clean state changes this by at most $0.01$. The learner's advantage is therefore a smooth function of the observed displacement beyond first order, consistent with the second-order term of the expansion, not a clean-state effect. $|\DGR|$ is the first-order part of this structure. It stays explanatory and needs no fitting, and it reports the direction and sign of the response, which the learner does not. Supplementary Table~S3 gives the full comparison.

\section{Clean-Side Approximation}
\label{sec:clean}

$\DGR$ uses the observed degraded displacement, so an obvious question is whether the response can be anticipated from the clean operating point. We test a clean-side approximation that uses no degraded data: the gradient at the clean point, multiplied by the mean displacement estimated on the other half of the cohort (cross-fitted). We also test the scale $\sqrt{\gradV^{\top}\hat\Sigma\gradV}$ of the response, where $\hat\Sigma$ is the covariance of the displacement on the other half.

The result is asymmetric across axes. On the video axis the approximation carries no usable information: the out-of-sample $R^2$ of the scale predictor is $0.02$--$0.08$, and the sign of the predicted response agrees with the observed sign at roughly chance level ($0.48$--$0.60$). On the audio axis a consistent direction is present across samples: at the strongest noise the correlation between the predicted and the observed response is $0.80$ on MSR-VTT ($0.42$ on DiDeMo), the sign agreement is $0.76$ ($0.70$), and the scale predictor reaches an out-of-sample $R^2$ of $0.60$ ($0.29$). The asymmetry matches the first-order error pattern: audio displacements have a stable direction that is predictable from the clean state, whereas blur displacements do not. Consequently $\DGR$ is an explanatory quantity, and its predictive counterpart is available for audio only.

\section{Gain Normalization Is Not a Removable Nuisance}
\label{sec:boundary}

The decomposition suggests a minimal intervention: divide the score by the local gain to remove operating-point dependence,
\begin{equation}
\begin{aligned}
\widetilde V(\Rel)&=\frac{V(\Rel)}{g_V(\Rel)+\varepsilon},\\
\varepsilon&=0.1\cdot\operatorname{median}(g_V).
\end{aligned}
\label{eq:norm}
\end{equation}
The candidate and its two gates were specified before the audit was executed. The \emph{plausibility gate} requires that the normalized response, $|\Delta V|/(g_V+\varepsilon)$, be better predicted from $D$ than $|\Delta V|$ itself, with a bootstrap interval excluding zero in at least three of the four original cells (the gate was specified before the Charades analysis existed). The \emph{clean-ordering gate} requires that $\widetilde V$ preserve the clean top-1 ordering of queries with keep rate at least $0.98$.

Under the corrected $D$ the plausibility gate is not met. Two cells show an interval excluding zero in favour of normalization: MSR-VTT video ($+0.069$, interval $[0.046,0.087]$) and DiDeMo video ($+0.104$, $[0.067,0.140]$). MSR-VTT audio is not distinguishable from zero ($-0.013$, $[-0.029,0.005]$), and DiDeMo audio is significantly worse ($-0.085$, $[-0.106,-0.057]$). The clean-ordering gate fails on both datasets: the top-1 keep rate is $0.9749$ on MSR-VTT and $0.9459$ on DiDeMo, below $0.98$, although the median Spearman correlation is high ($0.992$ and $0.988$). The same audit on Charades, added as a check outside the gate, gives $\Delta$OOS of $+0.055$ on video and $-0.017$ on audio, and a top-1 keep rate of $0.898$, so the clean-order gate fails there as well. An anisotropic covariance variant performs worse, with top-1 keep rates of $0.494$ and $0.391$. Its failure is in clean ordering, which is the binding gate on both datasets; the predictability gate alone does not separate the two variants.

The tested family therefore does not satisfy either pre-specified gate, and we do not change $\varepsilon$, learn a correction, or test post hoc variants. The result does not show that every normalization fails, nor that the local gain is unimportant: the gain carries partial signal on video (Section~\ref{sec:proxies}). It shows that the most direct gain-removal family of the decomposition does not remove the heterogeneity and distorts clean ordering. The gain is explanatory structure, not a removable nuisance under this intervention.

\begin{figure*}[t]
\centering
\includegraphics[width=\textwidth]{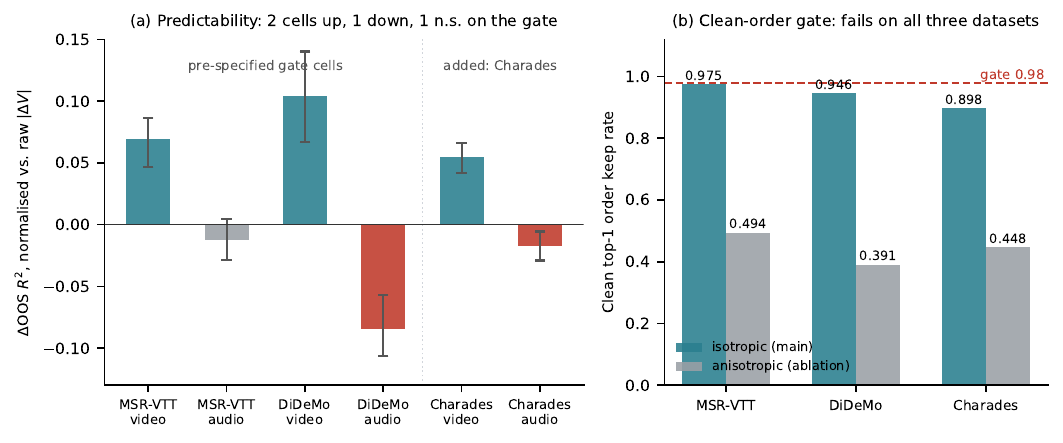}
\caption{Gain-normalization audit at the strongest level of each cell. The pre-specified gate uses the four original cells; Charades is shown as an added check. (a) Change in pooled out-of-sample $R^2$ with bootstrap interval, for predicting the normalized response $|\Delta V|/(g_V+\varepsilon)$ from $D$ compared with $|\Delta V|$ from $D$. The gate needs at least three of the four original cells with a positive interval; it has two. (b) Clean top-1 keep rate against the $0.98$ threshold; the isotropic candidate fails on both datasets and the anisotropic ablation fails severely.}
\label{fig:boundary}
\end{figure*}

\section{Consequence for Text-to-Video Retrieval}
\label{sec:retrieval}
The repository scores text-to-video retrieval with the negative Euclidean volume, $-V$, so that a matched video has the smaller volume. We use this convention, without the interaction re-ranking stage, on the 878-video cohort. Queries are clean text. The gallery is degraded in one modality: every candidate video is blurred ($\sigma=4$), or every candidate audio track receives noise at $0$~dB. The clean component places the correct video first for $42.9$\% of queries ($70.4$\% within five, $79.8$\% within ten). Video blur lowers top-1 accuracy to $40.0$\%. Audio noise gives $45.3$\%; this change is within the sampling uncertainty of 878 queries and we do not interpret it.

Rank changes are determined by pairwise gaps. For each query we take the ground-truth video and its ten highest-scoring clean competitors, and count a pair as flipped when its order reverses under degradation. Flips occur for $10.5$\% of pairs under video blur and $15.2$\% under audio noise. Supplementary Table~S5 reports the area under the ROC curve for three predictors of a flip. The clean margin alone gives $0.85$ and $0.81$, since small margins flip more often. The direction-free worst-case change $\|\nabla V\|D$ for both candidates gives $0.89$ and $0.68$. The first-order change $\DGR$ for both candidates, with the margin subtracted, gives $0.998$ and $0.986$. The predicted gap change also tracks the observed gap change, with Spearman correlation $0.997$ (video) and $0.968$ (audio).

This is an explanatory test with clear limits. The directional predictor uses the observed degraded displacement, so it is not a deployment-time predictor. Because the flip criterion includes the clean margin, a high AUC partly reflects how accurately the first-order term reproduces the gap change, which the Spearman figures measure directly. The test therefore establishes which rank reversals the first-order term accounts for, not a practical retrieval improvement. It covers the Euclidean component and one retrieval direction in the main text; the reverse direction and the full metrics are in Supplementary Table~S5.

All embeddings in this paper, including the clean embeddings used for retrieval, come from one end-to-end run in one software environment (Supplementary Section~S12). The retrieval experiment therefore uses the same clean embeddings as the main tables.

\section{Discussion}
\label{sec:discussion}

\subsection{What the Results Establish}
Four observations form a coherent chain. First, magnitude organizes the global spread of the response but not its ordering among samples of equal magnitude. Second, encoder-side signals and the clean operating point do not explain the within-magnitude ordering, and the direction-free product with magnitude remains clearly below the directional term. Third, the first-order directional term explains the within-magnitude ordering and the sign of the response across three datasets, two perturbation families, and all tested severities. Fourth, the clean-side gain is a partial signal on video only, and gain normalization fails its gates. Fifth, in text-to-video retrieval the same first-order term accounts for which rank orders reverse under gallery degradation. The audio--video asymmetry, where direction is predictable from the clean state for audio and not for video, is a separate finding that delimits the mechanism.

\subsection{What the Results Do Not Establish}
The analysis does not establish causality: the degradations are applied experimentally, but the explanatory claims are associational descriptions of a response mechanism. The directional term is not a deployment-time predictor, since it uses the observed displacement. It is not a theorem for all multimodal representations, and the evidence covers MSR-VTT, DiDeMo, and Charades, Gaussian blur, white noise, and the tested severities. The negative audit does not show that every normalization fails, only the tested family. Nothing here shows that HyperGRAM \cite{na2026hypergram} or Gramian volumes \cite{cicchetti2025gram} are flawed as objectives; the results characterize how the score responds to degradation, and Section~\ref{sec:retrieval} examines one retrieval protocol only.

\subsection{Implications and Limitations}
The results suggest that future geometric multimodal methods may need to consider local directional structure rather than treating perturbation severity as a scalar. For evaluation, studies of geometric scores may benefit from reporting displacement direction statistics alongside severity, and from matched-magnitude analyses rather than global correlations alone. For robustness-oriented design, rescaling a geometric score by a scalar gain may be mismatched to a mechanism in which direction carries the dominant within-magnitude heterogeneity.

The limitations are concrete. (1)~Three datasets (MSR-VTT, DiDeMo, Charades) and two perturbation families at the tested severities; ActivityNet \cite{cabahelbron2015activitynet} would be a natural external test. (2)~One Gramian-volume geometry \cite{cicchetti2025gram,na2026hypergram}; other geometric scores may respond differently. (3)~The first-order approximation is less accurate for strong audio noise, with median relative error of roughly $18$--$41$\%. (4)~The directional term is explanatory and uses the observed displacement. (5)~The clean-side approximation is informative for audio but not for video. (6)~The proxy analyses are restricted to MSR-VTT. (7)~The matching window was changed from an absolute to a relative definition after we found the absolute window not scale-invariant, and both are reported. (8)~The retrieval experiment uses the Euclidean component, one protocol, and the observed displacement. (9)~Absolute values depend on the software environment, in particular the audio decoding stack; all reported runs use one environment, and an earlier run with stored embeddings shifted some audio values by a few sign flips (Supplementary Section~S12). (10)~Charades has one description per video and quieter, sparser audio; its SNR is defined on the whole clip while the encoder sees a 10-second slice, and its audio gap to the learner is the largest we observe (Supplementary Section~S13).

\section*{Data and Code Availability}
The analysis code, the frozen cohort identifier lists, and the per-sample analysis outputs are publicly available at \url{https://github.com/VincentLuoMust/DGR}. The analysis uses frozen cohorts, fixed seeds, shared fold partitions within each cell and severity, and decision rules fixed before the gain-normalization audit was run. Charades was obtained from a public Kaggle mirror; Supplementary Section~S13 records the source and its checks. Cohort construction, perturbation definitions, and the numerical equivalence between the diagnostic and deployment geometry are described in the supplementary material.

\section*{Conflict of Interest}
The authors declare that they have no conflict of interest.

\section{Conclusion}
\label{sec:conclusion}
We studied how a multimodal Gramian volume responds when one modality is degraded. Across three video--text--audio datasets, two degradation families, and all tested severities, relational displacement magnitude explains a substantial share of the pooled response but does not order responses of equal magnitude. The first-order directional term, the projection of the displacement onto the local volume gradient, explains the within-magnitude ordering and the sign of the response, and outperforms direction-free combinations of gain and magnitude as well as encoder-side signals. A flexible learner given the same observed inputs is comparable overall and better on audio; its advantage is a smooth function of the observed displacement beyond first order, and in text-to-video retrieval the term accounts for which rank orders reverse. A clean-side approximation is informative for audio but not for video. The pre-specified gain normalization fails its gates, so the local gain remains explanatory structure rather than a calibration knob. Understanding and exploiting this directional structure, without assuming it can be divided out, is left for future work.

\bibliographystyle{IEEEtran}
\bibliography{refs_tmm}

@inproceedings{xu2016msrvtt,
  author    = {Xu, Jun and Mei, Tao and Yao, Ting and Rui, Yong},
  title     = {{MSR-VTT}: A Large Video Description Dataset for Bridging Video and Language},
  booktitle = {Proceedings of the IEEE Conference on Computer Vision and Pattern Recognition (CVPR)},
  year      = {2016},
  pages     = {5288--5296},
  doi       = {10.1109/CVPR.2016.571}
}

@inproceedings{hendricks2017didemo,
  author    = {Hendricks, Lisa Anne and Wang, Oliver and Shechtman, Eli and Sivic, Josef and Darrell, Trevor and Russell, Bryan},
  title     = {Localizing Moments in Video With Natural Language},
  booktitle = {Proceedings of the IEEE International Conference on Computer Vision (ICCV)},
  year      = {2017},
  pages     = {5803--5812}
}

@inproceedings{radford2021clip,
  author    = {Radford, Alec and Kim, Jong Wook and Hallacy, Chris and Ramesh, Aditya and Goh, Gabriel and Agarwal, Sandhini and Sastry, Girish and Askell, Amanda and Mishkin, Pamela and Clark, Jack and Krueger, Gretchen and Sutskever, Ilya},
  title     = {Learning Transferable Visual Models From Natural Language Supervision},
  booktitle = {Proceedings of the 38th International Conference on Machine Learning},
  volume    = {139},
  pages     = {8748--8763},
  year      = {2021},
  publisher = {PMLR}
}

@inproceedings{akbari2021vatt,
  author    = {Akbari, Hassan and Yuan, Liangzhe and Qian, Rui and Chuang, Wei-Hong and Chang, Shih-Fu and Cui, Yin and Gong, Boqing},
  title     = {{VATT}: Transformers for Multimodal Self-Supervised Learning from Raw Video, Audio and Text},
  booktitle = {Advances in Neural Information Processing Systems},
  volume    = {34},
  year      = {2021}
}

@inproceedings{girdhar2023imagebind,
  author    = {Girdhar, Rohit and El-Nouby, Alaaeldin and Liu, Zhuang and Singh, Mannat and Alwala, Kalyan Vasudev and Joulin, Armand and Misra, Ishan},
  title     = {{ImageBind}: One Embedding Space To Bind Them All},
  booktitle = {Proceedings of the IEEE/CVF Conference on Computer Vision and Pattern Recognition (CVPR)},
  pages     = {15180--15190},
  year      = {2023},
  doi       = {10.1109/CVPR52729.2023.01457}
}

@inproceedings{zhu2024languagebind,
  author    = {Zhu, Bin and Lin, Bin and Ning, Munan and Yan, Yang and Cui, Jiaxi and Wang, HongFa and Pang, Yatian and Jiang, Wenhao and Zhang, Junwu and Li, Zongwei and Zhang, Cai and Li, Zhifeng and Liu, Wei and Yuan, Li},
  title     = {{LanguageBind}: Extending Video-Language Pretraining to N-Modality by Language-Based Semantic Alignment},
  booktitle = {International Conference on Learning Representations (ICLR)},
  year      = {2024}
}

@inproceedings{chen2023vast,
  author    = {Chen, Sihan and Li, Handong and Wang, Qunbo and Zhao, Zijia and Sun, Mingzhen and Zhu, Xinxin and Liu, Jing},
  title     = {{VAST}: A Vision-Audio-Subtitle-Text Omni-Modality Foundation Model and Dataset},
  booktitle = {Advances in Neural Information Processing Systems},
  volume    = {36},
  pages     = {72842--72866},
  year      = {2023}
}

@inproceedings{cicchetti2025gram,
  author    = {Cicchetti, Giordano and Grassucci, Eleonora and Sigillo, Luigi and Comminiello, Danilo},
  title     = {Gramian Multimodal Representation Learning and Alignment},
  booktitle = {International Conference on Learning Representations (ICLR)},
  year      = {2025}
}

@inproceedings{na2026hypergram,
  author    = {Na, Saiyang and Jiang, Feng and Zhou, Qifeng and Zhong, Wenliang and Dang, Thao M. and Guo, Yuzhi and Ma, Hehuan and Li, Chunyuan and An, Weizhi and Huang, Junzhou},
  title     = {Hyperbolic Gramian Volumes for Multimodal Alignment},
  booktitle = {Proceedings of the IEEE/CVF Conference on Computer Vision and Pattern Recognition (CVPR)},
  pages     = {37756--37765},
  year      = {2026}
}

@inproceedings{nickel2017poincare,
  author    = {Nickel, Maximilian and Kiela, Douwe},
  title     = {Poincar{\'e} Embeddings for Learning Hierarchical Representations},
  booktitle = {Advances in Neural Information Processing Systems},
  volume    = {30},
  pages     = {6338--6347},
  year      = {2017}
}

@inproceedings{ganea2018hyperbolic,
  author    = {Ganea, Octavian-Eugen and B{\'e}cigneul, Gary and Hofmann, Thomas},
  title     = {Hyperbolic Neural Networks},
  booktitle = {Advances in Neural Information Processing Systems},
  volume    = {31},
  year      = {2018}
}

@inproceedings{desai2023meru,
  author    = {Desai, Karan and Nickel, Maximilian and Rajpurohit, Tanmay and Johnson, Justin and Vedantam, Ramakrishna},
  title     = {Hyperbolic Image-Text Representations},
  booktitle = {Proceedings of the 40th International Conference on Machine Learning},
  volume    = {202},
  pages     = {7694--7731},
  year      = {2023},
  publisher = {PMLR}
}

@article{baltrusaitis2019multimodal,
  author  = {Baltru{\v{s}}aitis, Tadas and Ahuja, Chaitanya and Morency, Louis-Philippe},
  title   = {Multimodal Machine Learning: A Survey and Taxonomy},
  journal = {IEEE Transactions on Pattern Analysis and Machine Intelligence},
  volume  = {41},
  number  = {2},
  pages   = {423--443},
  year    = {2019},
  doi     = {10.1109/TPAMI.2018.2798607}
}

@inproceedings{wang2020gradientblending,
  author    = {Wang, Weiyao and Tran, Du and Feiszli, Matt},
  title     = {What Makes Training Multi-Modal Classification Networks Hard?},
  booktitle = {Proceedings of the IEEE/CVF Conference on Computer Vision and Pattern Recognition (CVPR)},
  pages     = {12695--12705},
  year      = {2020},
  doi       = {10.1109/CVPR42600.2020.01271}
}

@inproceedings{ma2021smil,
  author    = {Ma, Mengmeng and Ren, Jian and Zhao, Long and Tulyakov, Sergey and Wu, Cathy and Peng, Xi},
  title     = {{SMIL}: Multimodal Learning with Severely Missing Modality},
  booktitle = {Proceedings of the AAAI Conference on Artificial Intelligence},
  volume    = {35},
  number    = {3},
  pages     = {2302--2310},
  year      = {2021},
  doi       = {10.1609/aaai.v35i3.16330}
}

@inproceedings{han2021trusted,
  author    = {Han, Zongbo and Zhang, Changqing and Fu, Huazhu and Zhou, Joey Tianyi},
  title     = {Trusted Multi-View Classification},
  booktitle = {International Conference on Learning Representations (ICLR)},
  year      = {2021}
}

@inproceedings{szegedy2014intriguing,
  author    = {Szegedy, Christian and Zaremba, Wojciech and Sutskever, Ilya and Bruna, Joan and Erhan, Dumitru and Goodfellow, Ian and Fergus, Rob},
  title     = {Intriguing Properties of Neural Networks},
  booktitle = {International Conference on Learning Representations (ICLR)},
  year      = {2014}
}

@inproceedings{goodfellow2015explaining,
  author    = {Goodfellow, Ian J. and Shlens, Jonathon and Szegedy, Christian},
  title     = {Explaining and Harnessing Adversarial Examples},
  booktitle = {International Conference on Learning Representations (ICLR)},
  year      = {2015}
}

@inproceedings{madry2018towards,
  author    = {Madry, Aleksander and Makelov, Aleksandar and Schmidt, Ludwig and Tsipras, Dimitris and Vladu, Adrian},
  title     = {Towards Deep Learning Models Resistant to Adversarial Attacks},
  booktitle = {International Conference on Learning Representations (ICLR)},
  year      = {2018}
}

@inproceedings{jia2021align,
  author    = {Jia, Chao and Yang, Yinfei and Xia, Ye and Chen, Yi-Ting and Parekh, Zarana and Pham, Hieu and Le, Quoc V. and Sung, Yun-Hsuan and Li, Zhen and Duerig, Tom},
  title     = {Scaling Up Visual and Vision-Language Representation Learning With Noisy Text Supervision},
  booktitle = {Proceedings of the 38th International Conference on Machine Learning},
  volume    = {139},
  pages     = {4904--4916},
  year      = {2021},
  publisher = {PMLR}
}

@article{peng2022hyperbolic,
  author  = {Peng, Wei and Varanka, Tuomas and Mostafa, Abdelrahman and Shi, Henglin and Zhao, Guoying},
  title   = {Hyperbolic Deep Neural Networks: A Survey},
  journal = {IEEE Transactions on Pattern Analysis and Machine Intelligence},
  volume  = {44},
  number  = {12},
  pages   = {10023--10044},
  year    = {2022},
  doi     = {10.1109/TPAMI.2021.3136921}
}

@inproceedings{musgrave2020metric,
  author    = {Musgrave, Kevin and Belongie, Serge and Lim, Ser-Nam},
  title     = {A Metric Learning Reality Check},
  booktitle = {Proceedings of the European Conference on Computer Vision (ECCV)},
  pages     = {681--699},
  year      = {2020}
}

@inproceedings{hendrycks2019corruptions,
  author    = {Hendrycks, Dan and Dietterich, Thomas},
  title     = {Benchmarking Neural Network Robustness to Common Corruptions and Perturbations},
  booktitle = {International Conference on Learning Representations (ICLR)},
  year      = {2019}
}

@inproceedings{moosavi2016deepfool,
  author    = {Moosavi-Dezfooli, Seyed-Mohsen and Fawzi, Alhussein and Frossard, Pascal},
  title     = {{DeepFool}: A Simple and Accurate Method to Fool Deep Neural Networks},
  booktitle = {Proceedings of the IEEE Conference on Computer Vision and Pattern Recognition (CVPR)},
  pages     = {2157--2163},
  year      = {2016}
}

@inproceedings{koh2017influence,
  author    = {Koh, Pang Wei and Liang, Percy},
  title     = {Understanding Black-Box Predictions via Influence Functions},
  booktitle = {Proceedings of the 34th International Conference on Machine Learning},
  volume    = {70},
  pages     = {1885--1894},
  year      = {2017},
  publisher = {PMLR}
}

@inproceedings{simonyan2014saliency,
  author    = {Simonyan, Karen and Vedaldi, Andrea and Zisserman, Andrew},
  title     = {Deep Inside Convolutional Networks: Visualising Image Classification Models and Saliency Maps},
  booktitle = {ICLR Workshop Track},
  year      = {2014}
}

@inproceedings{zhang2023provable,
  author    = {Zhang, Qingyang and Wu, Haitao and Zhang, Changqing and Hu, Qinghua and Fu, Huazhu and Zhou, Joey Tianyi and Peng, Xi},
  title     = {Provable Dynamic Fusion for Low-Quality Multimodal Data},
  booktitle = {Proceedings of the 40th International Conference on Machine Learning},
  volume    = {202},
  pages     = {41753--41769},
  year      = {2023},
  publisher = {PMLR}
}

@inproceedings{wang2023shaspec,
  author    = {Wang, Huaibo and Chen, Yanhuang and Ma, Chao and Avery, Jeova and Hull, Lansing and Carneiro, Gustavo},
  title     = {Multi-Modal Learning with Missing Modality via Shared-Specific Feature Modelling},
  booktitle = {Proceedings of the IEEE/CVF Conference on Computer Vision and Pattern Recognition (CVPR)},
  pages     = {15878--15887},
  year      = {2023}
}

@article{luo2022clip4clip,
  author  = {Luo, Huaishao and Ji, Lei and Zhong, Ming and Chen, Yang and Lei, Wen and Duan, Nan and Li, Tianrui},
  title   = {{CLIP4Clip}: An Empirical Study of {CLIP} for End to End Video Clip Retrieval and Captioning},
  journal = {Neurocomputing},
  volume  = {508},
  pages   = {293--304},
  year    = {2022},
  doi     = {10.1016/j.neucom.2022.07.028}
}

@inproceedings{cabahelbron2015activitynet,
  author    = {Caba Heilbron, Fabian and Escorcia, Victor and Ghanem, Bernard and Niebles, Juan Carlos},
  title     = {{ActivityNet}: A Large-Scale Video Benchmark for Human Activity Understanding},
  booktitle = {Proceedings of the IEEE Conference on Computer Vision and Pattern Recognition (CVPR)},
  pages     = {961--970},
  year      = {2015}
}

@article{pineau2021reproducibility,
  author  = {Pineau, Joelle and Vincent-Lamarre, Philippe and Sinha, Koustuv and Larivi{\`e}re, Vincent and Beygelzimer, Alina and d'Alch{\'e}-Buc, Florence and Fox, Emily and Larochelle, Hugo},
  title   = {Improving Reproducibility in Machine Learning Research: A Report from the {NeurIPS} 2019 Reproducibility Program},
  journal = {Journal of Machine Learning Research},
  volume  = {22},
  number  = {164},
  pages   = {1--20},
  year    = {2021}
}

@book{efron1993bootstrap,
  author    = {Efron, Bradley and Tibshirani, Robert J.},
  title     = {An Introduction to the Bootstrap},
  publisher = {Chapman \& Hall/CRC},
  year      = {1993}
}

@article{cameron2015cluster,
  author  = {Cameron, A. Colin and Miller, Douglas L.},
  title   = {A Practitioner's Guide to Cluster-Robust Inference},
  journal = {Journal of Human Resources},
  volume  = {50},
  number  = {2},
  pages   = {317--373},
  year    = {2015},
  doi     = {10.3368/jhr.50.2.317}
}

@inproceedings{poppi2025hyperbolic,
  author    = {Poppi, Samuele and Kasarla, Tejaswi and Mettes, Pascal and Baraldi, Lorenzo and Cucchiara, Rita},
  title     = {Hyperbolic Safety-Aware Vision-Language Models},
  booktitle = {Proceedings of the IEEE/CVF Conference on Computer Vision and Pattern Recognition (CVPR)},
  pages     = {4222--4232},
  year      = {2025}
}

@inproceedings{lee2023prompting,
  author    = {Lee, Yu-Lun and Tsai, Yi-Hsuan and Chiu, Wei-Chen and Lee, Chen-Yu},
  title     = {Multimodal Prompting with Missing Modalities for Visual Recognition},
  booktitle = {Proceedings of the IEEE/CVF Conference on Computer Vision and Pattern Recognition (CVPR)},
  pages     = {14943--14952},
  year      = {2023}
}

@inproceedings{sigurdsson2016charades,
  title={Hollywood in Homes: Crowdsourcing Data Collection for Activity Understanding},
  author={Sigurdsson, Gunnar A. and Varol, G{\"u}l and Wang, Xiaolong and Farhadi, Ali and Laptev, Ivan and Gupta, Abhinav},
  booktitle={European Conference on Computer Vision (ECCV)},
  year={2016}
}

@misc{luo2026arxiv,
  author    = {Luo, Yongsheng and He, Wengan and Li, Yu and Wu, Rouying and Lv, Wei},
  title     = {Beyond Perturbation Magnitude: Direction-Dependent Responses in Multimodal Geometric Representations},
  howpublished = {arXiv:2610.08533},
  year      = {2026},
  note      = {Preprint, versions v1 and v2}
}

\end{document}